\documentclass[10pt, a4paper]{article}
\usepackage{lrec2022} 
\usepackage{multibib}
\newcites{languageresource}{Language Resources}
\usepackage{graphicx}
\usepackage{tabularx}
\usepackage{soul}
\usepackage{titlesec}
\titleformat{\section}{\normalfont\large\bfseries\center}{\thesection.}{1em}{}
\titleformat{\subsection}{\normalfont\SmallTitleFont\bfseries\raggedright}{\thesubsection.}{1em}{}
\titleformat{\subsubsection}{\normalfont\normalsize\bfseries\raggedright}{\thesubsubsection.}{1em}{}
\renewcommand\thesection{\arabic{section}}
\renewcommand\thesubsection{\thesection.\arabic{subsection}}
\renewcommand\thesubsubsection{\thesubsection.\arabic{subsubsection}}

\usepackage{epstopdf}
\usepackage[utf8]{inputenc}

\usepackage{hyperref}
\usepackage{xstring}

\usepackage{color}

\usepackage{arabtex}
\usepackage{utf8}
\setcode{utf8}

\title{Harnessing Multilingual Resources to Question Answering in Arabic}

\name{Khalid Alnajjar and Mika Hämäläinen}

\address{Rootroo Ltd \\
         Helsinki, Finland \\
         \{firstname but don't spam\}@rootroo.com\\}

\abstract{
The goal of the paper is to predict answers to questions given a passage of Qur'an. The answers are always found in the passage, so the task of the model is to predict where an answer starts and where it ends. As the initial data set is rather small for training, we make use of multilingual BERT so that we can augment the training data by using data available for languages other than Arabic. Furthermore, we crawl a large Arabic corpus that is domain specific to religious discourse. Our approach consists of two steps, first we train a BERT model to predict a set of possible answers in a passage. Finally, we use another BERT based model to rank the candidate answers produced by the first BERT model.
 \\ \newline }

\begin{document}

\maketitleabstract

\section{Introduction}

Question answering is natural language understanding problem that has received a fair share of attention in the past \cite{nishida-etal-2019-answering,ruckle-etal-2020-multicqa,asai-choi-2021-challenges}. There are several datasets available for the task \citelanguageresource{rajpurkar2018know,artetxe2020cross,lewis2020mlqa}. These datasets cover a very different domain than the one we are interested in this paper, namely the holy script Qur'an.

Qur'anic Arabic itself has also received its share of NLP interest \cite{sharaf-atwell-2012-qurana,dukes-habash-2010-morphological,alsaleh-etal-2021-quranic}. There is even an earlier question answering system for Qur'an \cite{abdelnasser-etal-2014-al}. Other historical Arabic texts have also received some research interest \cite{belinkov-etal-2016-shamela,majadly-sagi-2021-dynamic,alnajjar2020automated}.

In this paper, we describe our work on the Qur'an QA 2022 shared task data. The problem the QRCD (Qur'anic Reading Comprehension Dataset) dataset \citelanguageresource{malhas2020ayatec} is built to solve is to predict an answer to a question given a passage in the Qur'an. The answer is within the passage, so the task for our model is to find where the answer starts and ends in the given passage.

The problem is a challenging one for several reasons. Firstly, Arabic language has greater degree of ambiguity in written form due to the fact that most of the diacritics are left out in writing. This means that several words that are pronounced different become homographs and have an identical written form. This ambiguity is not a characteristic of the language itself but much rather a result of the orthographic conventions. This ambiguity causes challenges not only in the dataset we are using but also in any pretrained Arabic language models.

Secondly, the publicly released part of the QRCD dataset is relatively small, consisting of only 710 samples in the training data and 109 samples in the development data. For this reason, we experiment with multilingual models and training data to alleviate this under-resourced scenario.

Thirdly, Arabic is a language with multiple dialects that are vastly different from each other. There is dedicated NLP research for several subdialects such as Tunisian \cite{ben-abdallah-etal-2020-text}, Palestinian \cite{jarrar-etal-2014-building}, Gulf \cite{adouane-johansson-2016-gulf} and Egyptian Arabic \cite{habash-etal-2012-morphological}. These are very different from the Qur'anic Arabic we are focusing on, but they will be present in any large scale language model trained for Arabic on online corpora. For this reason, we needed to ensure that the model we use in this paper is trained exclusively on Modern Standard Arabic as it is the closest contemporary variant of the language to Qur'anic Arabic and there are no language models available for classical Arabic.


\section{Dataset}
The QRCD Qur'an question answering dataset consists of 1,337 question-passage-answer triplets, which is split into training (65\%), development (10\%), and test (25\%) sets. As the amount of training data is small (i.e., 710 and 109 for training and validation, respectively), we leverage multilingual and crosslingual resources for question answering tasks while ensuring that the model is exposed to and aware of Islamic concepts.

To do so, we crawl multiple Islamic websites related to Tafseer (explanations of the Qur'an) and Fatwas (i.e., rulings or interpretations based on Islamic law for a given query) to build an Islamic-specific corpus. Table~\ref{tab:crawled-data} lists the web-sources we crawled and how many pages were retrieved per source. We crawled all the websites using Scrapy\footnote{https://github.com/scrapy/scrapy} except for quran-tafseer.com for which we used Pytafseer\footnote{https://github.com/Quran-Tafseer/pytafseer}. Additionally, we add the entire Qur'an to the dataset. We use a precleaned version of Qur'an that is stored in the UTF-8 format\footnote{https://github.com/aliftype/quran-data}.

\begin{table}[]
\centering
\begin{tabular}{|l|l|}
\hline
\textbf{Source}            & \textbf{N}     \\ \hline
aliftaa.jo        & 907   \\ \hline
binbaz.org.sa     & 23556 \\ \hline
islamqa.info      & 14950 \\ \hline
islamway.net      & 30978 \\ \hline
islamweb.net      & 30691 \\ \hline
quran-tafseer.com & 49888 \\ \hline
\end{tabular}%
\caption{The number of pages crawled per source in our Islam-specific corpus}
\label{tab:crawled-data}
\end{table}

In terms of question answering datasets, we adapt existing general (i.e., not Qur'an or Islamic related) QA datasets for other tasks. Namely, we use the Stanford Question Answering Dataset (SQuAD)~\citelanguageresource{rajpurkar2018know} MultiLingual Question Answering (MLQA)~\citelanguageresource{lewis2020mlqa} Cross-lingual Question Answering Dataset (XQuAD)~\citelanguageresource{artetxe2020cross}. Out of these MLQA and XQuAD also have Arabic data, which is beneficial when training the model. These datasets follow a slightly different format than the QRCD in terms of JSON. However, they are developed for the exact same task of predicting an answer in a text given a question. Therefore, they can be directly employed as additional training data without the need of reframing the problem.

\section{Approach}
In this section, we describe the process of building our artificial neural network model in detail. We base our model on the multilingual BERT \citelanguageresource{devlin-etal-2019-bert} that is trained on Wikipedia in multiple languages such as English, Spanish and Modern Standard Arabic. The intuition behind utilizing this model instead of an Arabic BERT model such as AraBERT \cite{antoun-etal-2020-arabert} is that there are more question answering dataset available in English than in Arabic; hence, our model would have a better representation for answering general questions related to a given context.

The additional perk of using multilingual BERT is that we know that it has been trained on Modern Standard Arabic as opposed to several dialects that may have been present in the training data of Arabic specific BERT models that have been trained on online data such as the Oscar corpus \cite{2022arXiv220106642A}. Qur'anic Arabic is closer to Modern Standard Arabic than any of the other Arabic dialects, which means that the model will have less noise coming from multiple different dialects as noise can have undesirable effects on the final results (see \cite{makela2020wrangling}). We also train the model using the Qur'an question answering dataset provided for the shared task to tailor its knowledge to the goal of the shared task. The following subsections elucidate on each of the aforementioned steps along with any preprocessing and postprocessing phases.

\subsection{Domain adaptation}

As the model we are basing our work on (i.e., multilingual BERT) is trained on a generic encyclopedia corpus (Wikipedia) and has little exposure to Islamic and Qur'anic concepts, we continue training the multilingual BERT model to adapt it to the domain of the task here. In our previous research \cite{hamalainen-etal-2021-never,hamalainen2021detecting}, we have found that BERT based models tend to work better if their training data has had text of a similar domain as the downstream task the model is fine-tuned for. Therefore, we believe that domain adaptation is beneficial in this case as well.

We convert the crawled data into a textual corpus, which we clean from non-Arabic text and remove any Arabic diacritics or punctuation using UralicNLP \cite{hamalainen2019uralicnlp}. In the case of Fatwas, we format the text as question first followed by the answer provided by the Mufti, and for Tafseer, we add the context (i.e., passage/verse) prior to the question. Qur'an data is added as it is. The textual corpus is then split into 80\% training and 20\% validation. We train the base model for the task of Masked Language Modeling and Next Sentence Prediction for 3 full epochs on our entire textual corpus.

\begin{table}[]
\centering
\resizebox{0.45\textwidth}{!}{%
\begin{tabular}{|l|l|c|c|c|c|}
\hline
\textbf{Question}      & \textbf{EN}   & \textbf{N} & \textbf{min} & \textbf{avg} & \textbf{max} \\ \hline
\RL{متى, كم, في + كم}         & When/How much & 56         & 1            & 10           & 55           \\ \hline
\RL{على + من, من, ضد + من, مع}  & Who           & 255        & 1            & 6            & 58           \\ \hline
\RL{ماهي, ماذا, ما, ب + ماذا} & What          & 387        & 1            & 8            & 225          \\ \hline
\RL{ب + أي, بأي, أي}          & Where         & 1          & 12           & 12           & 12           \\ \hline
\RL{ل + ماذا, لماذا, ف + لماذا} & Why           & 100        & 1            & 10           & 32           \\ \hline
\RL{هل}                     & Is it/Does    & 174        & 1            & 12           & 62           \\ \hline
\RL{أين}                    & Where         & 2          & 1            & 2            & 3            \\ \hline
\RL{كيف}                    & How           & 14         & 3            & 11           & 27           \\ \hline
\end{tabular}%
}
\caption{The different question types, their frequency and statistics on the length of their corresponding answers (in tokens) in the training and development dataset}
\label{tab:question-types}
\end{table}

\begin{table*}[]
\centering
\resizebox{\textwidth}{!}{%
\begin{tabular}{|l|ll|c|c|c|}
\hline
\textbf{\#} & \multicolumn{1}{l|}{\textbf{Model}} & \textbf{Training Data} & \textbf{pRR} & \textbf{Exact Match} & \textbf{F1@1} \\ \hline
1 & \multicolumn{1}{l|}{Base} & Quran QA training & 0.347 & 0.009 & 0.311 \\ \hline
2 & \multicolumn{1}{l|}{Base} & Quran QA training + SQuAD + MLQA + XQuAD & 0.351 & 0.092 & 0.288 \\ \hline
3 & \multicolumn{1}{l|}{Model 2} & Quran QA training & 0.373 & 0.028 & 0.345 \\ \hline
4 & \multicolumn{1}{l|}{Model 2} & \begin{tabular}[c]{@{}l@{}}Quran QA training + Quran QA development\end{tabular} & 0.540 & 0.092 & 0.526 \\ \hline
5 & \multicolumn{2}{l|}{Model 4 + post-processing} & 0.700 & \textbf{0.358} & \textbf{0.688} \\ \hline
\textbf{6} & \multicolumn{2}{l|}{\textbf{Model 4 + post-processing + recommendation}} & \textbf{0.704} & \textbf{0.358} & \textbf{0.688} \\ \hline
\textbf{7} & \multicolumn{2}{l|}{\textbf{Model 3.2 + Model 4 + post-processing + recommendation}} & 0.648 & 0.211 & 0.639 \\ \hline
\end{tabular}%
}
\caption{Experimental settings and their performances }
\label{tab:different-data-settings}
\end{table*}

\subsection{Fine-tuning for Question Answering}

Here, we fine-tune our BERT model for the task of Qur'an question answering. The data used for fine-tuning comes both from the QRCD dataset itself and the additional QA datasets (MLQA, SQuAD and XQuAD). The purpose of the other QA datsets is to make the model learn better the task of question-answering and make it learn to use multilingual information better for this task. In essence, this should improve the results for Arabic even though a majority of the training data was in a different language.

We append a fully-connected dense layer that accepts the BERT's hidden layers and outputs two vectors of predictions. These vectors are predictions for the start and end positions of the predicted answer in the context. We use the Adam algorithm~\cite{Adam} with decoupled weight decay regularization~\cite{AdamW} to optimize the parameters of the model, with cross entropy loss as the loss function.

\subsection{Predicting Answers and Post-processing them}

When inferring answers, we clamp the results to the number of tokens in the context and we ignore any special tokens prior to applying the softmax function on the predicted positions.

We apply a post-processing step on the top $N$ generated answers. This goal of this step is to 1) ensure that the length of the answer corresponds to the type of question that is being asked and 2) eliminate overlapping predictions.

Different types of questions require answers of different lengths; for instance, answers to ``who''-questions will most of the time be shorter (typically a single word is sufficient to answer it) than answers to ``why''-questions (requires further elaboration consisting of several tokens). For this reason, we apply some data analysis on the Quran QA dataset to find out what the questions are that are present in the dataset and what the  minimum, average and maximum lengths of the answers are. We use Farasa segmenter \cite{abdelali2016farasa} to process the data for this analysis.

As a given type of questions might be expressed in different ways, we have applied some manual clustering to group all asked questions in the dataset into 8 types based on the interrogative pronouns used. Table~\ref{tab:question-types} presents the question types, how many times they were present in the dataset and statistics on their answers. All predicted answers that are smaller than the average answer length are extended to either the nearest full-stop that marks the end of the verse or the average length, whichever is shorter.

When predicting multiple answers for a question, it is commonly the case that the model would predict overlapping answers. In such cases, we merge them together by taking the smaller starting position and maximum end position. 

\subsection{Similarity Recommendation}

We noticed during our observations of the Qur'an QA dataset that some questions are detailed and elaborate further on what is being sought as an answer. Such elaborations would indicate that the most semantically similar verse to the question is probably the answer to it. For this reason, we use the second version of AraBERT Large~\cite{antoun-etal-2020-arabert} to extract the features of the question and all verses in the given passage. Thereafter, we find the most similar verse to the question by applying the cosine similarity on the extracted features. The most similar verse is appended to the list of answers if it was not predicted already by the model and there are less than 5 predicted answers.

\begin{table}[]
\centering
\resizebox{0.45\textwidth}{!}{%
\begin{tabular}{|l|c|c|c|}
\hline
\textbf{Models}                                                                             & \textbf{pRR}   & \textbf{Exact Match} & \textbf{F1@1}    \\ \hline
KUISAIL's base BERT~\cite{safaya-etal-2020-kuisail}                                                                    & 0.286          & \textbf{0.037}       & 0.236          \\ \hline
KUISAIL's large BERT~\cite{safaya-etal-2020-kuisail}                                                                   & 0.330          & \textbf{0.037}       & 0.278          \\ \hline
CAMeLBERT Quarter~\cite{inoue-etal-2021-interplay} & 0.274          & 0.018                & 0.223          \\ \hline
Our BERT model                                                                       & \textbf{0.347} & 0.009                & \textbf{0.311} \\ \hline
\end{tabular}%
}
\caption{Different BERT models trained and evaluated solely on the Quran QA dataset}
\label{tab:vanilla-bert}
\end{table}

\begin{table}[]
\centering
\begin{tabular}{|l|c|c|c|}
\hline
      & \multicolumn{1}{l|}{\textbf{pRR}} & \multicolumn{1}{l|}{\textbf{Exact match}} & \multicolumn{1}{l|}{\textbf{F1@1}} \\ \hline
Run 6 & 0.392                    & 0.113                            & 0.354                     \\ \hline
Run 7 & 0.409                    & 0.092                            & 0.364                     \\ \hline
\end{tabular}%

\caption{Results on the test set}
\label{tab:fi-anal-results}
\end{table}

\begin{table*}[ht]
\centering
\resizebox{\textwidth}{!}{%
\begin{tabular}{|l|l|l|l|l|}
\hline
\textbf{\#} & \textbf{Context} & \textbf{Question} & \textbf{Gold Answers} & \textbf{Predicted Answers} \\ \hline
1 & \begin{tabular}[c]{@{}l@{}}\RL{وجعلنا الليل والنهار آيتين فمحونا آية الليل وجعلنا آية النهار}\\  \RL{مبصرة لتبتغوا فضلا من ربكم ولتعلموا عدد السنين والحساب}\\  \RL{وكل شيء فصلناه تفصيلا. وكل إنسان ألزمناه طائره في عنقه}\\  \RL{ونخرج له يوم القيامة كتابا يلقاه منشورا. اقرأ كتابك}\\  \RL{كفى بنفسك اليوم عليك حسيبا. من اهتدى فإنما يهتدي لنفسه}\\  \RL{ومن ضل فإنما يضل عليها ولا تزر وازرة وزر أخرى وما كنا معذبين}\\ \RL{ حتى نبعث رسولا. وإذا أردنا أن نهلك قرية أمرنا مترفيها ففسقوا}\\ \RL{ فيها فحق عليها القول فدمرناها تدميرا. وكم أهلكنا من القرون}\\ \RL{ من بعد نوح وكفى بربك بذنوب عباده خبيرا بصيرا.}\end{tabular} & \RL{هل سمح الإسلام بحرية الاعتقاد بالدخول إلى الإسلام؟} & \begin{tabular}[c]{@{}l@{}}\RL{من اهتدى فإنما يهتدي لنفسه ومن ضل فإنما}\\  \RL{يضل عليها ولا تزر وازرة وزر أخرى وما كنا }\\ \RL{معذبين حتى نبعث رسولا}\end{tabular} & \begin{tabular}[c]{@{}l@{}}1. \RL{وكل إنسان ألزمناه طائره في عنقه}\\ 2. \RL{من اهتدى فإنما يهتدي لنفسه}\end{tabular} \\ \hline
2 & \begin{tabular}[c]{@{}l@{}}\RL{وقل الحق من ربكم فمن شاء فليؤمن ومن شاء فليكفر}\\  \RL{إنا أعتدنا للظالمين نارا أحاط بهم سرادقها وإن يستغيثوا يغاثوا}\\ \RL{ بماء كالمهل يشوي الوجوه بئس الشراب وساءت مرتفقا. }\\ \RL{إن الذين آمنوا وعملوا الصالحات إنا لا نضيع أجر من أحسن عملا. }\\ \RL{أولئك لهم جنات عدن تجري من تحتهم الأنهار يحلون فيها من}\\  \RL{أساور من ذهب ويلبسون ثيابا خضرا من سندس وإستبرق}\\  \RL{متكئين فيها على الأرائك نعم الثواب وحسنت مرتفقا}\end{tabular} & \RL{هل سمح الإسلام بحرية الاعتقاد بالدخول إلى الإسلام؟} & \RL{من شاء فليؤمن ومن شاء فليكفر} & \begin{tabular}[c]{@{}l@{}}1. \RL{فمن شاء فليؤمن ومن شاء فليكفر}\\ 2. \RL{إن الذين آمنوا وعملوا الصالحات}\\ \RL{ إنا لا نضيع أجر من أحسن عملا}\end{tabular} \\ \hline
3 & \begin{tabular}[c]{@{}l@{}}\RL{لقد كان لكم في رسول الله أسوة حسنة لمن كان يرجو الله}\\  \RL{واليوم الآخر وذكر الله كثيرا. ولما رأى المؤمنون الأحزاب قالوا}\\  \RL{هذا ما وعدنا الله ورسوله وصدق الله ورسوله وما زادهم إلا إيمانا}\\  \RL{وتسليما. من المؤمنين رجال صدقوا ما عاهدوا الله عليه فمنهم}\\  \RL{من قضى نحبه ومنهم من ينتظر وما بدلوا تبديلا. ليجزي الله الصادقين}\\  \RL{بصدقهم ويعذب المنافقين إن شاء أو يتوب عليهم إن الله كان }\\ \RL{غفورا رحيما.}\end{tabular} & \RL{هل هناك إسلام بدون الحديث الشريف؟} & \RL{لقد كان لكم في رسول الله أسوة حسنة} & \begin{tabular}[c]{@{}l@{}}\RL{لقد كان لكم في رسول الله أسوة حسنة لمن}\\  \RL{كان يرجو الله واليوم الآخر وذكر الله كثيرا}\end{tabular} \\ \hline
4 & \begin{tabular}[c]{@{}l@{}}\RL{كما أرسلنا فيكم رسولا منكم يتلو عليكم آياتنا ويزكيكم ويعلمكم}\\  \RL{الكتاب والحكمة ويعلمكم ما لم تكونوا تعلمون. فاذكروني أذكركم}\\  \RL{واشكروا لي ولا تكفرون. يا أيها الذين آمنوا استعينوا بالصبر والصلاة}\\ \RL{ إن الله مع الصابرين.}\end{tabular} & \RL{هل هناك إسلام بدون الحديث الشريف؟} & \begin{tabular}[c]{@{}l@{}}1. \RL{أرسلنا فيكم رسولا منكم يتلو عليكم}\\ \RL{آياتنا ويزكيكم ويعلمكم الكتاب والحكمة}\\ 2. \RL{يعلمكم ما لم تكونوا تعلمون}\end{tabular} & \begin{tabular}[c]{@{}l@{}}1. \RL{أرسلنا فيكم رسولا منكم يتلو عليكم آياتنا ويز}\\ \RL{كيكم ويعلمكم الكتاب والحكمة ويعلمكم ما لم}\\  \RL{تكونوا تعلمون}\\ 2. \RL{الذين آمنوا استعينوا بالصبر والصلاةفاذكروني}\\  \RL{أذكركم واشكروا لي ولا تكفرون}\end{tabular} \\ \hline
5 & \begin{tabular}[c]{@{}l@{}}\RL{وإن يريدوا أن يخدعوك فإن حسبك الله هو الذي أيدك بنصره}\\ \RL{ وبالمؤمنين. وألف بين قلوبهم لو أنفقت ما في الأرض جميعا ما ألفت}\\ \RL{ بين قلوبهم ولكن الله ألف بينهم إنه عزيز حكيم. يا أيها النبي حسبك}\\ \RL{ الله ومن اتبعك من المؤمنين. يا أيها النبي حرض المؤمنين على}\\  \RL{القتال إن يكن منكم عشرون صابرون يغلبوا مائتين وإن يكن منكم}\\ \RL{ مائة يغلبوا ألفا من الذين كفروا بأنهم قوم لا يفقهون. الآن خفف}\\  \RL{الله عنكم وعلم أن فيكم ضعفا فإن يكن منكم مائة صابرة يغلبوا}\\  \RL{مائتين وإن يكن منكم ألف يغلبوا ألفين بإذن الله والله مع الصابرين.}\end{tabular} & \RL{ضد من فُرض الجهاد؟} & \RL{الذين كفروا} & \begin{tabular}[c]{@{}l@{}}1. \RL{أيها النبي حرض المؤمنين}\\ 2. \RL{وألف بين قلوبهم لو أنفقت ما في}\\  \RL{الأرض جميعا ما ألفت بين قلوبهم ولكن}\\  \RL{الله ألف بينهم إنه عزيز حكيم}\end{tabular} \\ \hline
\end{tabular}%
}
\caption{Examples of input contexts and questions to the system along with the gold answers and what the model has predicted}
\label{tab:examples}
\end{table*}

\section{Results and Evaluation}

We experiment with different models and techniques for predicting answers. First, we test out different BERT models and compare them to our custom model. Secondly, we investigate the effects of fine-tuning our model with different training and validation question and answering datasets along with the Quran QA dataset. Lastly, we assess the benefits of post-processing the predictions and including the most similar verse to the question as an answer. In our tests, we evaluate the models based on the metrics that are considered in the shared task, namely partial Reciprocal Rank (pRR)~\cite{ppr}, exact match and F1~\citelanguageresource{rajpurkar-etal-2016-squad}.

Table~\ref{tab:vanilla-bert} shows the different BERT models that we have tested out using only the Quran QA dataset, where we use the train and validation step during the training phase, and test the models on the development split. Comparing KUISAIL's base and large models suggest that bigger models improve the performance of the model. However, larger models require a longer time to train, and for this reason we opted for using a base model. Despite using a smaller multilingual model as a base model, adapting it to the domain of this task has clearly improved the quality of its predictions. All results presented after this point use our custom BERT model.

In Table \ref{tab:fi-anal-results} we can see the results of our two systems when comparing to the official test set of the shared task. When we compared our system on a question level to the median values across all the submissions to the task, we found that over half of the time our best system achieves better scores than the median value. However, our best model had the highest possible score among all submissions around 15\% of the time. Interestingly, our worst model had the highest possible score among all the submissions 17.6\% of the time despite having poorer overall performance.

Table~\ref{tab:different-data-settings} lists the different settings we experimented with and their evaluation results on the development split. All the models had the Quran QA dataset as the validation dataset during the training phase, and they have been trained for 3 epochs. By comparing the first and second settings, we see that including other question and answering datasets during the training phase improves the predictions. Fine-tuning the model that has been exposed to other question and answering datasets further using Quran QA dataset only outperforms using Quran QA dataset solely, which demonstrates the great importance of utilizing relevant linguistic resources in other languages and applying domain adaptation. In the 4th experimental setting we include the Quran QA development split in the training dataset to cover as many cases as possible given that the amount of training data is very small; despite it being a non-recommended practice.

Our experiments point out that post-processing the predicted answers to ensure that they are of an adequate length based on the question type and that no overlapping answers have been predicted boosts the results from pRR of 0.54 to 0.7. Including the most similar verse to the question as a possible answer raises the results by a bit but it does not affect them negatively.

From our observations of answers predicted by model \#6 and \#7 is that sometimes they would predict different answers where one of them is correct. To benefit from both of the models and include their variations in the answers, we consider answers produced by them and remove any overlapping answers during the post-processing phase. We have submitted two runs to the shared task, which are experiments number 6 and 7.

Looking at examples of generated answers by our models in Table~\ref{tab:examples} illustrates cases where the predictions have been fully accurate by predicting an exact match (e.g. example \#2) and partially correct (e.g., examples \#1, \#3 and \#4). For the partially correct predictions, the model either predicts lengthier or shorter answers which could be due to the post-processing phase. However, we find that the length of gold answers is subjective. The fifth case is an example of wrong predictions; nonetheless, the top prediction is includes two nouns and the question is asking for people (whom) which tells us that the model made its best guess given the context and it was very close.

\section{Conclusions}

In conclusion, we have embraced multilingual models, and question and answering resources to build a question answering model for Qur'an. Our results indicate that applying domain adaptation and fine-tuning the model with relevant data sets increases the performance of the models, especially in the case of limited training data like this one.

As the models predict the start and end positions of the answer in the context, it is very likely that the predictions are off by few tokens. Post-processing the predictions to correspond to the expected answer per question type and merging any overlapping cases had a huge boost on the quality of predictions.


\section{Bibliographical References}\label{reference}

\bibliographystyle{lrec2022-bib}
\bibliography{lrec2022-example}

\section{Language Resource References}
\label{lr:ref}
\bibliographystylelanguageresource{lrec2022-bib}
\bibliographylanguageresource{languageresource}

\end{document}